# Eight Maximal Tractable Subclasses of Allen's Algebra with Metric Time

**Thomas Drakengren**                                    THODR@IDA.LIU.SE
**Peter Jonsson**                                        PETEJ@IDA.LIU.SE
*Department of Computer and Information Science, Linköping University*
*S-581 83 Linköping, Sweden*

## Abstract

This paper combines two important directions of research in temporal resoning: that of finding maximal tractable subclasses of Allen's interval algebra, and that of reasoning with metric temporal information. Eight new maximal tractable subclasses of Allen's interval algebra are presented, some of them subsuming previously reported tractable algebras. The algebras allow for metric temporal constraints on interval starting or ending points, using the recent framework of *Horn DLRs*. Two of the algebras can express the notion of *sequentiality* between intervals, being the first such algebras admitting both qualitative and metric time.

## 1. Introduction

Reasoning about temporal knowledge abounds in artificial intelligence applications and other areas, such as planning (Allen, 1991), natural language understanding (Song and Cohen, 1988) and molecular biology (Benzer, 1959; Golumbic and Shamir, 1993). However, since even the restricted problem of reasoning with pure qualitative time in Allen's interval algebra (Allen, 1983) is NP-complete (Vilain et al., 1989), research has focused on identifying classes of problems where reasoning is tractable (Drakengren and Jonsson, 1996; Golumbic and Shamir, 1993; Kautz and Ladkin, 1991; Nebel and Bürckert, 1995; van Beek and Cohen, 1990; van Beek, 1989; van Beek, 1992).

Until recently, approaches have mostly been either metric or qualitative, with a few exceptions (Kautz and Ladkin, 1991; Meiri, 1991; Gerevini et al., 1993). However, the approach of Jonsson and Bäckström (1996) (also developed independently by Koubarakis, 1996) manages to unify almost every approach to tractable reasoning about metric time, qualitative time and the integrated approaches in one framework, that of *Horn disjunctive linear relations* (Horn-DLRs), in which the reasoning problem can be solved in polynomial time. Since its expressiveness in terms of qualitative information subsumes that of the maximal tractable[1] ORD-Horn algebra (Nebel and Bürckert, 1995), it can be viewed as a maximal tractable subalgebra of Allen's interval algebra, provided with metric temporal information.

First, this paper continues the work on finding maximal tractable subalgebras of Allen's algebra started by Nebel and Bürckert (1995) and continued by Drakengren and Jonsson (1996), by identifying eight more maximal tractable subclasses of Allen's algebra. Second, we combine this with the work of Jonsson and Bäckström (1996), by providing the maximal algebras with metric temporal information in the form of Horn DLRs, whose expressiveness subsumes that of the ORD-Horn algebra. Further, the maximality result of these algebras settles the question of maximality of some algebras in Drakengren and Jonsson (1996), since these are included in the new algebras presented here. Two of the maximal algebras can express the notion of *sequentiality* between intervals[2], important for example in reasoning about action (Sandewall, 1994), where actions are typically assumed to occur in sequence. To our knowledge, these are the first such algebras in the literature that are also provided with metric temporal information. By the fact that we can also relate starting or ending points by Horn DLRs, we have a combination (with restrictions, of course) of the expressiveness of the ORD-Horn algebra, and that of sequentiality.

---

1. That is, it is tractable, and no other tractable algebra strictly contains it.

| Basic relation | | Example | Endpoints |
|---|---|---|---|
| $x$ before $y$ | $\prec$ | xxx | $x^+ < y^-$ |
| $y$ after $x$ | $\succ$ |    yyy | |
| $x$ meets $y$ | m | xxxx | $x^+ = y^-$ |
| $y$ met-by $x$ | m⌣ |    yyyy | |
| $x$ overlaps $y$ | o | xxxx | $x^- < y^- < x^+$, |
| $y$ overl.-by $x$ | o⌣ |   yyyy | $x^+ < y^+$ |
| $x$ during $y$ | d |  xxx | $x^- > y^-$, |
| $y$ includes $x$ | d⌣ | yyyyyyy | $x^+ < y^+$ |
| $x$ starts $y$ | s | xxx | $x^- = y^-$, |
| $y$ started by $x$ | s⌣ | yyyyyyy | $x^+ < y^+$ |
| $x$ finishes $y$ | f |     xxx | $x^+ = y^+$, |
| $y$ finished by $x$ | f⌣ | yyyyyyy | $x^- > y^-$ |
| $x$ equals $y$ | $\equiv$ | xxxx | $x^- = y^-$, |
| | | yyyy | $x^+ = y^+$ |

Table 1: The thirteen basic relations. The endpoint relations $x^- < x^+$ and $y^- < y^+$ that are valid for all relations have been omitted.

The structure of the paper is as follows. Section 2 introduces Allen's interval algebra, Section 3 defines the concepts needed to integrate qualitative and metric reasoning, together with the satisfiability algorithm, and Section 4 presents the new maximal tractable algebras, and how to provide these with metric temporal information. In Section 5 and Section 6, we prove that the algorithm is correct and that the algebras are maximal. Finally, Section 7 and Section 8 discuss and conclude the paper.

## 2. Allen's Interval Algebra

Allen's interval algebra (Allen, 1983) is based on the notion of *relations between pairs of intervals*. An interval $x$ is represented as a tuple $\langle x^-, x^+ \rangle$ of real numbers with $x^- < x^+$, denoting the left and right endpoints of the interval, respectively, and relations between intervals are composed as disjunctions of *basic interval relations*, which are those in Table 1. Such disjunctions are represented as *sets* of basic relations, but using a notation such that, for example, the disjunction of the basic intervals $\prec$, m and f$^\smile$ is written ($\prec$ m f$^\smile$). Thus, we have that ($\prec$ f$^\smile$) $\subseteq$ ($\prec$ m f$^\smile$). Sometimes the disjunction of *all* basic relations is written $\top$ and the empty relation is written $\bot$ (this is also used for relations between interval endpoints, denoting "always satisfiable" and "unsatisfiable", respectively). The algebra is provided with the operations of *converse, intersection and composition* on intervals, but we shall need only the converse operation explicitly. The converse operation takes an interval relation $i$ to its converse $i^\smile$, obtained by inverting each basic relation in $i$, i.e., exchanging $x$ and $y$ in the endpoint relations shown in Table 1.

---

2. That is, they contain the relations $(\prec)$, $(\succ)$, $(\prec \succ)$, $(\equiv \prec)$, $(\equiv \succ)$, $(\equiv \prec \succ)$, and the "unrelated" relation, and can thus express that "two intervals occur in sequence".

By the fact that there are thirteen basic relations, we get $2^{13} = 8192$ possible relations between intervals in the full algebra. We denote the set of all interval relations by $\mathcal{A}$. Subclasses of the full algebra are obtained by considering subsets of $\mathcal{A}$. There are $2^{8192} \approx 10^{2466}$ such subclasses. Classes that are closed under the operations of intersection, converse and composition are said to be *algebras*.

Although there are several computational problems associated with Allen's interval algebra, this paper focuses on the problem of *satisfiability (ISAT)* of a set of interval variables with relations between them, i.e., deciding whether there exists an assignment of intervals on the real line for the interval variables, such that all of the relations between the intervals are satisfied. We define this as follows.

**Definition 2.1 (*ISAT($\mathcal{I}$)*)**
Let $\mathcal{I}$ be a set of interval relations. An instance of *ISAT($\mathcal{I}$)* is a labelled directed graph $S = \langle V, E \rangle$, where the nodes in $V$ are interval variables and $E$ is a subset of $V \times \mathcal{I} \times V$. A labelled edge $\langle u, r, v \rangle \in E$ means that $u$ and $v$ are related by $r$.

A function $M$ taking an interval variable $v$ to its interval representation $M(v) = \langle x^-, x^+ \rangle$ with $x^-, x^+ \in \mathbb{R}$ and $x^- < x^+$, is said to be an *interpretation* of $S$.

An instance $S = \langle V, E \rangle$ is said to be *satisfiable* iff there exists an interpretation $M$ such that for each $\langle u, r, v \rangle \in E$, $M(u)rM(v)$ holds, i.e., the endpoint relations required by $r$ (see Table 1) are satisfied by the assignments of $u$ and $v$. Then $M$ is said to be a *model* of $S$.

We refer to the *size* of an instance $S$ as $|V| + |E|$. $\square$

For $\mathcal{A}$, we have the following result.

**Proposition 2.2** *ISAT($\mathcal{A}$)* is NP-complete.
**Proof:** See Vilain et al. (1989). $\square$

## 3. Qualitative and Metric Time

We first briefly recapitulate the Horn-DLR formalism of Jonsson and Bäckström (1996).

**Definition 3.1 (Linear relation, Disjunctive linear relation)**
Let $X = \{x_1, \ldots, x_n\}$ be a set of real-valued variables, and $\alpha, \beta$ linear polynomials (polynomials of degree one) over $X$ with rational coefficients. A *linear relation* over $X$ is a mathematical expression of the form $\alpha r \beta$, where $r \in \{<, \leq, =, \neq, \geq, >\}$. A *disjunctive linear relation* (DLR) over $X$ is a disjunction of one or more linear relations. A DLR is said to be *Horn* iff at most one of its disjuncts is not of the form $\alpha \neq \beta$.

The satisfiability problem for finite sets $H$ of Horn DLRs is denoted $\text{HORN DLR\textsc{sat}}(H)$, checking whether there exists an assignment $M$ of variables in $X$ to real numbers such that all DLRs in $H$ are satisfied in $M$. Such an $M$ is said to be a *model* of $H$. □

**Example 3.2**

$$x + 2y \leq 3z + 42.3$$

is a linear relation,

$$(x + 2y \leq 3z + 42.3) \vee (x > \frac{3}{12})$$

is a disjunctive linear relation, and

$$(x + 2y \leq 3z + 42.3) \vee (x \neq \frac{3}{12})$$

is a Horn disjunctive linear relation. □

**Proposition 3.3** There is a polynomial-time algorithm for $\text{HORN DLR\textsc{sat}}(H)$.
**Proof:** See Jonsson and Bäckström (1996) or Koubarakis (1996). □

In principle, the framework of DLRs makes it unnecessary to distinguish between qualitative and metric information. Nevertheless, when it comes to identifying tractable subclasses, the distinction is still convenient.

The Horn-DLR approach subsumes almost all previously known approaches to tractable metric and qualitative temporal reasoning, e.g. (Nebel and Bürckert, 1995; Koubarakis, 1992; Dechter et al., 1991; Meiri, 1991; Gerevini et al., 1993). It is worth mentioning that the maximal tractable algebras found by Drakengren and Jonsson (1996) cannot be expressed as Horn DLRs.

Although polynomial, the algorithm presented in Jonsson and Bäckström (1996) is quite expensive (it relies on a linear-programming algorithm) so when we have no need of specifying metric information, the following well-known subclass of the set of Horn DLRs will result in a lower-complexity algorithm.

**Definition 3.4 (The point algebra)**
The *point algebra* (Vilain, 1982) is the subclass of Horn DLRs consisting of the set of expressions $xRy$, where $x$ and $y$ are variables, and $R$ is one of the relations $<$, $\leq$, $=$, $\neq$, $\geq$ and $>$. □

The satisfiability problem for this subclass is denoted $\text{PA\textsc{sat}}(H)$, for a set $H$ of point algebra formulae.

**Proposition 3.5** $\text{PA\textsc{sat}}(H)$ is solvable in linear time in the size of $H$.
**Proof:** See Gerevini et al. (1993) (for practical purposes the algorithm of Delgrande and Gupta, 1996, could be preferred). □

Next, we define the problem of interest in this paper — the interval satisfiability problem with metric temporal information.

**Definition 3.6** (*M-ISAT(I)*)

Let $\langle V, E \rangle$ be an instance of *ISAT(I)* and $H$ a finite set of DLRs over the set $\{v^+, v^- \mid v \in V\}$ of variables, $v^-$ representing starting points and $v^+$ ending points of intervals $v$.

An instance of the problem of *interval satisfiability with metric information* for a set $I$ of interval relations, denoted *M-ISAT(I)*, is a tuple $Q = \langle V, E, H \rangle$.

An *interpretation* $M$ for $Q$ is an interpretation for $\langle V, E \rangle$. Since we now need to refer to starting and ending points of intervals, we extend the notation such that $M(v^-)$ obtains the starting point of the interval $M(v)$, and similarly for $M(v^+)$.

An instance $Q$ is said to be *satisfiable* iff there exists a model $M$ of $\langle V, E \rangle$ such that the DLRs in $H$ are satisfied, with values for all $v^-$ and $v^+$ by $M(v^-)$ and $M(v^+)$, respectively. □

Since every Allen interval relation can be expressed as a DLR (but not necessarily as a Horn DLR), we could instead have formulated the problem as a pure satisfiability problem of a set of DLRs, but since we are interested in the particular structure imposed on the problem by interval relations specifically we prefer this formulation.

Several concepts are needed in order to present the *starting and ending point algebras*, for which we shall provide polynomial-time algorithms. The curious reader might temporarily jump to Section 4 for the explicit presentation of the algebras which will be proved to be starting or ending points algebras.

The following definitions are needed to transfer information from interval relations to point relations.

**Definition 3.7** ($sprel(r)$, $eprel(r)$, $sprel^+(r)$, $eprel^-(r)$)

Take the relation $r \in \mathcal{A}$, let $u$ and $v$ be interval variables, and consider the instance $S$ of *ISAT(\{r\})* which relates $u$ and $v$ with the relation $r$ only. Define the relation $sprel(r)$ on real numbers to be the symbol for the implied relation between the starting points of $u$ and $v$. That is, for basic relations define

$$
\begin{aligned}
sprel(\equiv) &= \text{``=''} \\
sprel(\prec) &= \text{``<''} \\
sprel(\mathsf{d}) &= \text{``>''} \\
sprel(\mathsf{o}) &= \text{``<''} \\
sprel(\mathsf{m}) &= \text{``<''} \\
sprel(\mathsf{s}) &= \text{``=''} \\
sprel(\mathsf{f}) &= \text{``>''} \\
sprel(r^\smile) &= (sprel(r))^{-1},
\end{aligned}
$$

and for disjunctions $sprel(r)$ is the relation symbol corresponding to $\bigvee_{b \in r} sprel(b)$. For example, $sprel((\prec \succ)) = \text{``}\neq\text{''}$. Symmetrically, we define $eprel(r)$ to be the implied relation between ending points given $r$. Note that $sprel(r)$ and $eprel(r)$ have to be either of $<, \leq, =, \geq, >, \neq, \top$ or $\bot$.

Further, we define specialisations of these, by $sprel^+(r) = sprel(r \cap (\equiv \mathsf{f}\ \mathsf{f}^\smile))$ and $eprel^-(r) = eprel(r \cap (\equiv \mathsf{s}\ \mathsf{s}^\smile))$, i.e., the implied relations on starting (ending) points by $r$, given that the ending (starting) points are known to be equal. □

**Definition 3.8 (Explicit starting (ending) point relations)**
Let $\mathcal{I} \subseteq \mathcal{A}$, $Q = \langle V, E, H \rangle$ an instance of $M\text{-}ISAT(\mathcal{I})$, and construct the instance $Q' = \langle V, E, H' \rangle$ of $M\text{-}ISAT(\mathcal{I})$ by setting

$$H' = H \cup \{u^- \, sprel(r) v^- \mid \langle u, r, v \rangle \in E\}.$$

Then $Q'$ is said to be obtained from $Q$ by *making starting point relations explicit*. We denote this $Q'$ by $expl^-(Q)$.

Symmetrically, using *eprel* and ending points instead of *sprel* and starting points, $Q'$ is said to be obtained from $Q$ by *making ending points explicit*, denoted $expl^+(Q)$. $\square$

It is easy to see that only point algebra formulae are added to $H$.

Transferring information from interval relations to point relations does not change satisfiability, as expected:

**Proposition 3.9** Let $\mathcal{I} \subseteq \mathcal{A}$ and $Q$ an instance of $M\text{-}ISAT(\mathcal{I})$. Then $Q$ is satisfiable iff $expl^-(Q)$ is satisfiable iff $expl^+(Q)$ is satisfiable.
**Proof:** By the fact that the added starting and ending point relations are already guaranteed to hold in any model of $Q$. $\square$

We jump ahead by presenting a satisfiability algorithm (Algorithm 3.10) and briefly discuss the intuition behind it in order to indicate what kind of algebras it works for. This will hopefully make it easier to appreciate Definition 3.13.

First assume that $H$ only contains Horn DLRs, which only relate starting points of intervals. Line 1 makes the interval relations explicit as starting point relations and line 2 checks satisfiability of the resulting set of starting point relations. Lines 4 to 11 collect in $K$ the relations $u^- = v^-$, such that in any model these starting points have to be equal. In addition, $K$ forces all starting points to be distinct, that are not forced to be equal. It is clear that the equality formulae in $K$ do not affect satisfiability. However, it is less clear that the disequality formulae in $K$ cannot make the instance $Q'' = \langle V, E, H' \cup K \rangle$ unsatisfiable. This fact indeed follows from a property of Horn DLRs, which is proved in Theorem 5.9. At line 12, we know that there are no two models for $Q''$, where for some $u, v \in V$, $u^- = v^-$ in one model, and $u^- \neq v^-$ in the other model. This is the intuition behind Definition 3.11. Now, line 13 checks for satisfiability of the ending points of those intervals whose starting points have to be equal in any model. If the algorithm rejects at line 14, then the instance is obviously not satisfiable. Otherwise we need a condition on the algebra $\mathcal{I}$, corresponding to Definition 3.13, in order to guarantee satisfiability.

The formal machinery follows.

**Definition 3.11 (Starting (ending) point definite)**
Let $\mathcal{I} \subseteq \mathcal{A}$, and $Q = \langle V, E, H \rangle$ an instance of $M\text{-}ISAT(\mathcal{I})$. The instance $Q' = \langle V, E, H \cup H' \rangle$ of $M\text{-}ISAT(\mathcal{I})$ is said to be *starting point definite wrt. $Q$* iff there exists a function $f : E \to \{=, \neq\}$ such that $H' = \{u^- f(e) v^- \mid \langle u, r, v \rangle \in E\}$. We denote this relation by $def^-(Q, Q')$. This means that for each relation, either the starting points of related intervals are forced to be equal in all models, or they are forced to be distinct in all models. If for some $Q$ we have $def^-(Q, Q')$, then $Q'$ is said to be *starting point definite*.

Similarly, by exchanging starting and ending points, we get that $Q'$ is *ending point definite wrt. $Q$*, denoted $def^+(Q, Q')$. $\square$

**Algorithm 3.10** $(M_s\text{-}ISAT(\mathcal{I}))$

    **input** Instance $Q = \langle V, E, H \rangle$

1   $Q' = \langle V, E, H' \rangle \leftarrow expl^-(Q)$
2   **if** not HORNDLRSAT($H'$) **then**
3      **reject**
4   $K \leftarrow \emptyset$
5   **for** each $\langle u, r, v \rangle \in E$
6      **if** not HORNDLRSAT($H' \cup \{u^- \neq v^-\}$) **then**
7         $K \leftarrow K \cup \{u^- = v^-\}$
8      **else**
9         $K \leftarrow K \cup \{u^- \neq v^-\}$
10    **endif**
11  **endfor**
12  $P \leftarrow \{u^+ eprel^-(r)v^+ \mid \langle u, r, v \rangle \in E \wedge u^- = v^- \in H' \cup K\}$
13  **if** not PASAT($P$) **then**
14    **reject**
15  **accept**

$\square$

Line 13 of Algorithm 3.10 checks the following, as we shall see.

**Definition 3.12 (Locally satisfiable for starting (ending) points)**
Let $Q = \langle V, E, H \rangle$ be a starting point definite instance of $M\text{-}ISAT(\mathcal{I})$ for some $\mathcal{I} \subseteq \mathcal{A}$, and construct $Q' = \langle V, E', H \rangle$ such that $E' = \{\langle u, r, v \rangle \in E \mid u^- = v^- \in H\}$, i.e. by considering only relations which force the starting points to be equal. Now $Q$ is said to be *locally satisfiable for starting points* iff $Q'$ is satisfiable. A model satisfying $Q'$ is said to *locally satisfy $Q$ for starting points*.

    Similarly, exchanging starting and ending points, $Q$ is said to be *locally satisfiable for ending points* iff $Q'$ is satisfiable, and a model satisfying $Q'$ is said to *locally satisfy $Q$ for ending points*. $\square$

We now define the algebra for which Algorithm 3.10 solves satisfiability.

**Definition 3.13 (Starting (ending) point algebra)**
A subalgebra $\mathcal{I} \subseteq \mathcal{A}$ is said to be a *starting point algebra* iff for any instance $Q = \langle V, E, H \rangle$ of $M\text{-}ISAT(\mathcal{I})$, the following holds: for any $T = \langle V, E, H' \rangle$ such that $def^-(expl^-(Q), T)$, if $T$ is locally satisfiable for starting points, then $T$ is satisfiable.

    Symmetrically, exchanging ending points and starting points, we obtain an *ending point algebra*. $\square$

The satisfiability problems for these algebras are defined as follows.

**Definition 3.14** ($M_s\text{-}ISAT(\mathcal{I})$, $M_e\text{-}ISAT(\mathcal{I})$)

Let $\mathcal{I}$ be a starting point algebra. The *satisfiability problem for starting point algebras with metric information* is the set of instances $\langle V, E, H \rangle$ of $M\text{-}ISAT(\mathcal{I})$ where the DLRs of $H$ are restricted in two ways: first, $H$ may only contain Horn DLRs and second, $H$ may not contain any variables $v^+$, where $v \in V$, i.e., it may only relate starting points of intervals.

Symmetrically, by exchanging starting and ending points, we get the *satisfiability problem for ending point algebras with metric information*. □

## 4. Tractable Algebras

We now present the algebras which are starting (ending) point algebras.

**Definition 4.1 (The subclasses $S(b)$ and $E(b)$)**

Set $r_s = (\succ \ \mathsf{d} \ \mathsf{o}^\smile \ \mathsf{m}^\smile \ \mathsf{f})$, and $r_e = (\prec \ \mathsf{d} \ \mathsf{o} \ \mathsf{m} \ \mathsf{s})$. Note that $r_s$ contains all basic relations $b$ such that whenever $I b J$ for interval variables $I$, $J$, $I^- > J^-$ has to hold in any model and symmetrically, $r_e$ is equivalent to $I^+ < J^+$ holding in any model.

First, for $b \in \{\succ, \mathsf{d}, \mathsf{o}^\smile\}$, define $S(b)$ to be the set of relations $r$, such that either of the following holds:

$$
\begin{aligned}
(b \ b^\smile) \ &\subseteq \ r \\
(b) \ &\subseteq \ r \ \subseteq \ r_s \cup (\equiv \ \mathsf{s} \ \mathsf{s}^\smile) \\
(b^\smile) \ &\subseteq \ r \ \subseteq \ r_s{}^\smile \cup (\equiv \ \mathsf{s} \ \mathsf{s}^\smile) \\
r \ &\subseteq \ (\equiv \ \mathsf{s} \ \mathsf{s}^\smile).
\end{aligned}
$$

Then, by switching starting and ending points of intervals, for $b \in \{\prec, \mathsf{d}, \mathsf{o}\}$, $E(b)$ is defined to be the set of relations $r$, such that either of the following holds:

$$
\begin{aligned}
(b \ b^\smile) \ &\subseteq \ r \\
(b) \ &\subseteq \ r \ \subseteq \ r_e \cup (\equiv \ \mathsf{f} \ \mathsf{f}^\smile) \\
(b^\smile) \ &\subseteq \ r \ \subseteq \ r_e{}^\smile \cup (\equiv \ \mathsf{f} \ \mathsf{f}^\smile) \\
r \ &\subseteq \ (\equiv \ \mathsf{f} \ \mathsf{f}^\smile).
\end{aligned}
$$

□

**Definition 4.2 (The subclasses $S^*$ and $E^*$)**

Let $r_s$ and $r_e$ be as in Definition 4.1, and define $S^*$ to be the set of relations $r$, such that either of the following holds:

$$
\begin{aligned}
(\equiv \ \mathsf{f} \ \mathsf{f}^\smile) \ &\subseteq \ r \\
(\mathsf{f} \ \mathsf{f}^\smile) \ &\subseteq \ r \ \subseteq \ r_s \cup r_s{}^\smile \\
(\equiv \ \mathsf{f}) \ &\subseteq \ r \ \subseteq \ r_s \cup (\equiv \ \mathsf{s} \ \mathsf{s}^\smile) \\
(\equiv \ \mathsf{f}^\smile) \ &\subseteq \ r \ \subseteq \ r_s{}^\smile \cup (\equiv \ \mathsf{s} \ \mathsf{s}^\smile) \\
(\mathsf{f}) \ &\subseteq \ r \ \subseteq \ r_s \\
(\mathsf{f}^\smile) \ &\subseteq \ r \ \subseteq \ r_s{}^\smile \\
(\equiv) \ &\subseteq \ r \ \subseteq \ (\equiv \ \mathsf{s} \ \mathsf{s}^\smile) \\
r \ &= \ \bot
\end{aligned}
$$

Symmetrically, replacing f by s (and their inverses), ($\equiv$ s s$^\smile$) by ($\equiv$ f f$^\smile$), and $r_s$ by $r_e$, we get the subclass $E^*$. □

Files containing the algebras are supplied as an on-line appendix to this article.

It is easy to see that the algebras are all considerably larger than, for example, the ORD-Horn algebra, which contains 868 elements.

**Proposition 4.3** The six algebras $S(b)$ and $E(b)$ contain 2312 elements each and $S^*$ and $E^*$ contain 1445 elements each.
**Proof:** A straightforward combinatorial exercise according to the definitions. □

We also see that the $S(b)$ and $E(b)$ algebras each contain five basic relations, and that $S^*$ and $E^*$ contain three basic relations each. A subsumption result and a nonsubsumption result follow.

**Proposition 4.4** The twelve algebras presented by Drakengren and Jonsson (1996), which were not classified as maximal tractable, are each included in one of the algebras $S(b)$ and $E(b)$.
**Proof:** By simply checking inclusion from the definitions. □

**Proposition 4.5** In all of the algebras $S(b)$, $E(b)$, $S^*$ and $E^*$, there are relations which are not expressible by Horn DLRs alone.
**Proof:** It is easily verified that the point relations induced by the Allen relations ($\prec \succ$), (d d$^\smile$), (o o$^\smile$), ($\succ$ f$^\smile$) and ($\prec$ s) are not Horn DLRs. □

It was observed by Drakengren and Jonsson (1996) that the ORD-Horn algebra cannot express the notion of *sequentiality*, and thus since it is maximal tractable, we cannot add the relation ($\prec \succ$) to it without losing tractability. However, we can obtain a weaker yet useful result by the following observation: We know from the results of Jonsson and Bäckström (1996) that the expressivity of Horn DLRs subsumes that of the ORD-Horn algebra, by expressing the ORD-Horn relations as disjunctions of point relations in the starting and ending points of the intervals. Thus, since the satisfiability problem for starting point algebras (and ending point algebras, which follows by symmetry) allow arbitrary Horn DLRs relating starting points, we can convert any network expressed in the ORD-Horn algebra into an equivalent instance of $M_s$-$ISAT(\mathcal{I})$ for some of the tractable subclasses above, where only starting points of intervals are related. The additional expressivity of the starting point algebras can then be used to express e.g. sequentiality (using one of the algebras $S(\succ)$ or $E(\prec)$) or other relations between intervals.

Now it is time to verify that the presented algebras are indeed starting and ending point algebras, respectively. A few auxiliary definitions and results are needed.

**Definition 4.6 (Sign function)**
For $x \in \mathbb{R}$, let $sgn(x) \in \{-1, 0, 1\}$ be the *sign* of $x$, that is, if $x < 0$, then $sgn(x) = -1$, if $x = 0$ then $sgn(x) = 0$, and if $x > 0$, then $sgn(x) = 1$. □

**Lemma 4.7** Let $Q = \langle V, E, H \rangle$ be a starting point definite instance of $M_s$-$ISAT(\mathcal{I})$, which is locally satisfiable for starting points by some model $M$, and let $M'$ be an interpretation for $Q$ such that

- $\forall v \in V.M(v^-) = M'(v^-)$

- $\forall u, v \in V.M(u^-) = M(v^-) \rightarrow$
  $sgn(M(u^+) - M(v^+)) = sgn(M'(u^+) - M'(v^+))$,

which means that $M'$ may not differ from $M$ in the interpretation of starting points, and for ending points, any change is allowed, as long as their relative order for relations which have the same interpretations of starting points is the same. Then $M'$ also locally satisfies $Q$ for starting points.

**Proof:** Apart from checking the DLRs in $H$, local satisfiability for starting points checks only relations where the intervals they relate are forced to have the same starting point. Since $H$ does not relate ending points of intervals, the only thing that affects satisfiability of these relations is the relative order of ending points, given a fixed starting point. Since this order is the same, and $M$ and $M'$ coincide on starting points, the result follows. $\square$

**Lemma 4.8** Let $Q = \langle V, E, H \rangle$ be a starting point definite instance of $M_s\text{-}ISAT(\mathcal{I})$, where for no $\langle u, r, v \rangle \in E$, $r \cap (\equiv \ \mathsf{s}\ \mathsf{s}^\smile) \neq \emptyset$ and $r - (\equiv \ \mathsf{s}\ \mathsf{s}^\smile) \neq \emptyset$. Then $Q$ is satisfied by the model $M$ iff $M$ locally satisfies $Q$ for starting points and $M$ satisfies $\langle V, E' \rangle$, for $E' = \{\langle u, r, v \rangle \in E \mid r \cap (\equiv \ \mathsf{s}\ \mathsf{s}^\smile) = \emptyset\}$.

**Proof:**
$\Rightarrow$) Assuming that $M$ satisfies $Q$, the latter condition is a direct consequence of the definitions.
$\Leftarrow$) By the restriction on $Q$, satisfiability of every relation $r$ is checked by the two conditions together, and the satisfiability of $H$ is included in the local satisfiability condition. Thus $Q$ is satisfiable. $\square$

**Lemma 4.9** Let $Q = \langle V, E, H \rangle$ be an instance of $M\text{-}ISAT(\mathcal{I})$, and let $T = \langle V, E, H' \rangle$ be such that $def^-(Q, T)$. Construct $T' = \langle V, E', H' \rangle$ by setting

$$E' = \{\langle u, r \cap (\equiv \ \mathsf{s}\ \mathsf{s}^\smile), v \rangle \mid v_1^- = v_2^- \in H\} \cup$$
$$\{\langle u, r - (\equiv \ \mathsf{s}\ \mathsf{s}^\smile), v \rangle \mid v_1^- \neq v_2^- \in H\}$$

Now $T'$ is satisfiable iff $T$ is. The analogous result holds for ending points, when references to starting points are changed to ending points, and $(\equiv \ \mathsf{s}\ \mathsf{s}^\smile)$ is changed to $(\equiv \ \mathsf{f}\ \mathsf{f}^\smile)$.

**Proof:** Directly from the definitions. The restrictions imposed on the qualitative relations are already guaranteed to hold in any model, by the restrictions on $H'$. Note that $E'$ is well-defined since $def$ adds to $H$ either equality or inequality for all interval starting points.

The property of ending points follows by symmetry. $\square$

**Definition 4.10 (Absolute value)**
For $x \in \mathbb{R}$, denote by $abs(x)$ the absolute value of $x$, i.e. $x \cdot sgn(x)$. $\square$

The main results follow.

**Theorem 4.11** The algebras $S(b)$ are starting point algebras, and the algebras $E(b)$ are ending point algebras.

**Proof:** Let $Q = \langle V, E, H \rangle$ be an instance of $M_s\text{-}ISAT(S(b))$, and let $T = \langle V, E, H' \rangle$ with $def^-(expl^-(Q), T)$. By Lemma 4.9, we can assume that for every $\langle u, r, v \rangle \in E$, either $r \subseteq (\equiv \ \mathsf{s}\ \mathsf{s}^\smile)$ or $r \cap (\equiv \ \mathsf{s}\ \mathsf{s}^\smile) = \emptyset$, since $T$ is starting point definite. Thus, the only relations $r$ which are left are those satisfying

$$
\begin{aligned}
(b \; b^\smile) &\subseteq r \subseteq r_s \cup r_s{}^\smile \\
(b) &\subseteq r \subseteq r_s \\
(b^\smile) &\subseteq r \subseteq r_s{}^\smile \\
r &\subseteq (\equiv \mathsf{s}\,\mathsf{s}^\smile).
\end{aligned}
$$

So, suppose that $T$ is locally satisfied for starting points by a model $M$. Since $T$ has explicit starting points, all relations from line 4 are satisfied in $M$. Also, since $M$ imposes a certain order on starting points of intervals, we know that without loss of generality, relations $r$ on line 1 can be replaced by either $r \cap r_s$ or $r \cap r_s{}^\smile$, since $r_s$ and $r_s{}^\smile$ impose disjoint orderings on starting points. Thus, without loss of generality, we take $T$ only to contain relations from lines 2 and 4, inverting relations containing line 3 relations while changing their direction. If we could modify $M$ into an interpretation $M'$ such that the conditions of Lemma 4.7 are satisfied, and the relations of line 2 and their inverses are satisfied, then by Lemma 4.8, since these relations do not overlap with $(\equiv \mathsf{s}\,\mathsf{s}^\smile)$, $M'$ would be a model of $T$, and the result would follow. Indeed, we shall satisfy line 2 relations with the basic relation $b$ on every arc.

A few auxiliary definitions are needed for the construction. Define $MD(M)$ to be the least nonzero member of the set $\{abs(M(u^-) - M(v^-)) \mid u, v \in V\}$, i.e., the least nonzero distance between starting points in $M$. Given a $w \in V$, let $EP(w)$ be the set

$$
\{v \in V \mid M(v^-) = M(w^-)\},
$$

i.e., the set of intervals whose order of ending points has to be fixed, to maintain local satisfiability, by Lemma 4.7. Note that $EP(w)$ is uniquely determined by $M(w^-)$. Also, for a given $w$, let $n_w = |\{M(v^+) \mid v \in EP(w)\}|$ (the number of distinct ending points in $M$ within $w$'s "group"), and $f_w : EP(w) \to \{k \in \mathbb{N} \mid k < n_w\}$ the function uniquely determined by the ordering on ending points of $EP(w)$ in $M$, such that for every $u, v \in EP(w)$,

$$
sgn(M(u^+) - M(v^+)) = sgn(f_w(u) - f_w(v)).
$$

Note that for every $u, v \in EP(w)$, $f_u = f_v$.

Let $s = |\{M(v^-) \mid v \in V\}|$, i.e., the number of distinct starting points in $M$. Corresponding to $f_w$, define $i : V \to \{k \in \mathbb{N} \mid k < s\}$ to be the uniquely determined function such that for every $u, v \in V$,

$$
sgn(M(u^-) - M(v^-)) = sgn(i(u) - i(w)).
$$

We construct the interpretation $M'$ as follows, depending on $b$, and afterwards prove that it is a model of $T$. First, set $\epsilon = MD(M)$, and for all $v \in V$, set $M'(v^-) = M(v^-)$.

- Suppose that $b$ is $\succ$. For every $v \in V$, set

$$
M'(v^+) = M(v^-) + \frac{\epsilon}{4}(1 + \frac{f_v(v)}{n_v}).
$$

- Suppose that $b$ is $\mathsf{d}$. For every $v \in V$, set

$$
M'(v^+) = M(v^-) + 1 + 2\epsilon(s - i(v) - 1) + \frac{\epsilon}{2}(\frac{f_v(v)}{n_v} - 1).
$$

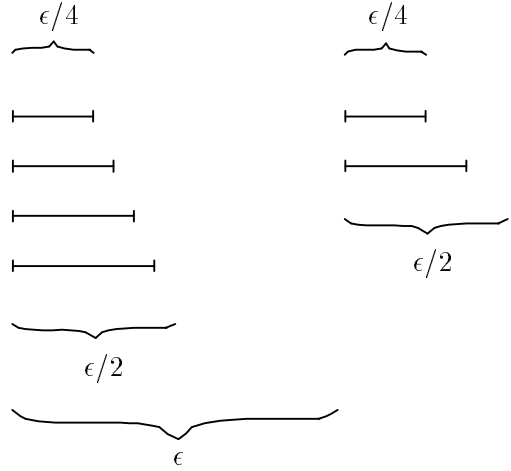

Figure 1: An example of the construction for the case when $b$ is $\succ$ in the proof of Theorem 4.11.

- Suppose that $b$ is $\mathsf{o}^{\smile}$. First let $w \in V$ be such that $i(w) = s - 1$, i.e., having the largest ending point, set $M'(w^+) = M(w^-) + 1$. For every $v \in V$, set

$$M'(v^+) = M(w^-) + \frac{i(v) + 1}{s} + \frac{1}{s}\left(\frac{f_v(v)}{n_v} - \frac{1}{2}\right).$$

Now, by construction, the models satisfy $b$ on every arc. Furthermore, the order of ending points within each $EP(v)$ is retained, and the orderings on starting points are identical in $M$ and $M'$. Thus $M'$ locally satisfies $T$ for starting points by Lemma 4.7, and by Lemma 4.8, $M'$ is a model of $T$.

The proof for $E(b)$ is symmetrical. $\square$

Fortunately, the next proof is much more convenient.

**Theorem 4.12** The algebra $S^*$ is a starting point algebra, and $E^*$ is an ending point algebra.

**Proof:** We prove only the $S^*$ case, since the $E^*$ case is symmetrical.

Precisely as in the proof of Theorem 4.11, we obtain an instance $T$ labelled only by relations $r$ satisfying

$$
\begin{aligned}
(\mathsf{f}\ \mathsf{f}^{\smile}) &\subseteq r \subseteq r_s \cup r_s{}^{\smile} \\
(\mathsf{f}) &\subseteq r \subseteq r_s \\
(\mathsf{f}^{\smile}) &\subseteq r \subseteq r_s{}^{\smile} \\
(\equiv) &\subseteq r \subseteq (\equiv\ \mathsf{s}\ \mathsf{s}^{\smile}) \\
r &= \perp.
\end{aligned}
$$

First assume that $T$ is locally satisfied for starting points by some model $M$. Since $M$ imposes a certain order on starting points of intervals, we know that relations $r$ on line 1 now can be assumed to be either $r \cap r_s$ or $r \cap r_s{}^{\smile}$, since relations in $r_s$ and $r_s{}^{\smile}$ impose disjoint orderings on starting points, as in the proof of Theorem 4.11. Thus, without loss of generality, we take $T$ to contain only relations from lines 2 and 4, inverting line 3 relations while changing their direction (relations from line 5 make every instance unsatisfiable). We shall show that $T$ is always satisfiable by constructing a model $M'$ of $T$. First set $M'(v^-) = M(v^-)$ for every $v \in V$. Let $x$ be maximal such that $x = M(v^-)$, i.e., $x$ is the largest starting point in $M$. Then set $M'(v^+) = x + 1$ for every $v \in V$. Obviously, for $u, v \in V$, whenever $M'(u^-) > M'(v^-)$, then $u(\mathbf{f})v$ is satisfied in $M'$. For line 4 relations $r$, when $urv$, it has to hold that $M'(u^-) = M'(v^-)$, and thus by construction $M'(u^+) = M'(v^+)$, and $r$ is satisfied. Thus $M'$ is a model of $T$ and the result follows. $\square$

It remains to show that Algorithm 3.10 correctly solves satisfiability for starting and ending point algebras.

## 5. Correctness of the Algorithm

Several concepts are needed for the correctness proof.

**Definition 5.1 (Convex combination, Convex)**
Given $x, y \in \mathbb{R}^n$, a *convex combination* of $x$ and $y$ is any $z \in \mathbb{R}^n$ of the form $\theta x + (1 - \theta)y$, where $0 \leq \theta \leq 1$.

A set $S \subseteq \mathbb{R}^n$ is said to be *convex* iff any convex combination of elements in $S$ is already in $S$. $\square$

**Definition 5.2 (Hyperplane)**
A *hyperplane* in $\mathbb{R}^n$ is a non-empty set defined as

$$\{x \in \mathbb{R}^n \mid \alpha(x) = b\},$$

for a linear polynomial $\alpha$ in $x_1, \ldots, x_n$, and $b \in \mathbb{R}$. $\square$

**Definition 5.3 (Almost convex)**
A set $S \subseteq \mathbb{R}^n$ is said to be *almost convex* if for any $x, y \in S$, at most finitely many of the convex combinations of $x$ and $y$ are not in $S$. $\square$

**Example 5.4** The set
$$S = \mathbb{R}^3 - \{(x, y, z) \mid x, y, z \in \mathbb{Z}\}$$

is almost convex. $\square$

**Proposition 5.5** Any intersection of finitely many almost convex sets is almost convex.
**Proof:** Let $S_1, \ldots, S_k$ be almost convex sets and let $S = S_1 \cap \ldots \cap S_k$. Take $x, y \in S$. Then $x, y \in S_i$ for every $1 \leq i \leq k$. Thus finitely many of the convex combinations of $x$ and $y$ are not in $S_i$. But since we intersect only a finite number of sets, only finitely many convex combinations are excluded from $S$, and the result follows. $\square$

The following is a generalisation of Lemma 13 of Jonsson and Bäckström (1996) (which is a simplified version of the proof by Lassez and McAloon, 1992), from convex to almost convex sets.

**Lemma 5.6** Let $S \subseteq \mathbb{R}^n$ be an almost convex set, and let $H_1, \ldots, H_k$ be hyperplanes. If $S \subseteq \bigcup_{i=1}^{k} H_i$, then there exists a $j$, $1 \leq j \leq k$ such that $S \subseteq H_j$.

**Proof:** Induction on $k$, letting $\mathcal{H}_k = \bigcup_{i=1}^{k} H_i$. For $k = 1$, the result is immediate, since $\mathcal{H}_1 = H_1$. Assuming that the result holds for $k$, we wish to show that if $S \subseteq \mathcal{H}_{k+1}$, then there exists a $j$, $1 \leq j \leq k + 1$ such that $S \subseteq H_j$. So, suppose $S \subseteq \mathcal{H}_{k+1}$, i.e., that $S \subseteq \mathcal{H}_k \cup H_{k+1}$. If $S \subseteq \mathcal{H}_k$, then the result follows by induction. Also if $S \subseteq H_{k+1}$, or if one of $\mathcal{H}_k$ or $H_{k+1}$ is included in the other, the result follows. Thus, suppose that $x, y \in S$, $x \in \mathcal{H}_k$, $y \in H_{k+1}$, satisfying $x \notin H_{k+1}$ and $y \notin \mathcal{H}_k$, which is the remaining case.

Consider the line segment $L$ adjoining $x$ and $y$, which is the convex combinations of $x$ and $y$. Every hyperplane either contains $L$ or intersects $L$ in at most one point. If some $H_i$ contains $L$, then $x, y \in H_i$, violating the choice of $x$ and $y$. Thus $L$ is intersected by hyperplanes $H_i$ in at most finitely many points, and those are the only members of $L$ which can be in $\mathcal{H}_{k+1}$. Since $S$ is almost convex, there are infinitely many points of $L$ which are in $S$ and are not members of any hyperplane $H_i$, contradicting that this remaining case could hold. The result follows. $\square$

One may note that this result holds even if $S$ satisfies the weaker property of containing infinitely many convex combinations of each pair of elements, but then the proof of Proposition 5.5 would not go through.

**Lemma 5.7** Let $\gamma$ be a satisfiable Horn DLR, and let

$$S(\gamma) = \{x \in \mathbb{R}^n \mid \gamma \text{ satisfied by } x\}$$

be the set of all solutions to $\gamma$. Then $S(\gamma)$ is almost convex.

**Proof:** By the definition of a Horn DLR, for some convex set $C$ and hyperplanes $H_i$, we have

$$S(\gamma) = C \cup \bigcup_{i=1}^{k} \overline{H_i} = C \cup \overline{\bigcap_{i=1}^{k} H_i}.$$

Take $x, y \in S(\gamma)$. If $x, y \in C$, then every convex combination of $x$ and $y$ is in $S(\gamma)$ by the convexity of $C$. If $y \notin C$ (and $x$ is either in $C$ or not), then $y \in \overline{H_l}$ for some $l$. Let $L$ be the line segment adjoining $x$ and $y$. The hyperplane $H_l$ either contains $L$ or intersects $L$ in at most one point. It cannot contain $L$, since $y \notin H_l$; thus $H_l$ intersects $L$ in at most one point, and the remaining points of $L$ are members of $\overline{H_l}$, and thus are members of $S(\gamma)$. $\square$

**Lemma 5.8** Let $H$ be a satisfiable set of Horn DLRs, and let

$$S(H) = \{x \in \mathbb{R}^n \mid \text{every } \gamma \in H \text{ is satisfied by } x\}$$

be the set of all solutions to $H$. Then $S(H)$ is almost convex.

**Proof:** Directly from Lemma 5.7 and Proposition 5.5. $\square$

The following is the key result for obtaining correctness of the algorithm.

**Theorem 5.9** Let $H$ be a satisfiable set of Horn DLRs, and let $x_1, x_2, \ldots, x_n$ be the variables used in $H$. Also define the set $\Gamma$ of Horn DLRs by

$$\Gamma = \{x_i \neq x_j \mid \{x_i \neq x_j\} \cup H \text{ is satisfiable}\}.$$

Then $H \cup \Gamma$ is satisfiable.

**Proof:** Suppose that $H \cup \Gamma$ is not satisfiable, although $H$ is satisfiable. Let $S$ be the set of solutions of $H$, as defined in Lemma 5.8. By definition, the set of solutions to a formula $x_i \neq x_j \in \Gamma$ is a complement of some hyperplane $H_{ij}$. Thus, the set of solutions to $H \cup \Gamma$ is $S - \bigcup_{i,j} H_{ij} = \emptyset$, equivalent to $S \subseteq \bigcup_{i,j} H_{i,j}$. By Lemma 5.8, $S$ is almost convex, and by Lemma 5.6, $S \subseteq H_{kl}$ for some hyperplane $H_{kl}$. But then $S \cap H_{kl} = \emptyset$, contrary to our assumption, and the result follows. $\square$

Thus, it is enough to check the formulae in $\Gamma$ *separately* for satisfiability. The correctness proof of the algorithm follows.

**Theorem 5.10** Algorithm 3.10 correctly solves satisfiability for starting point algebras. Symmetrically, by exchanging starting and ending points of Algorithm 3.10, it correctly solves satisfiability for ending point algebras.

**Proof:** Suppose that $Q$ is satisfiable. Then after line 1, $Q'$ is satisfiable, by Proposition 3.9. Thus we cannot get reject at line 3, since $H'$ has to be satisfied for $Q'$ to be satisfiable. Consider the value of $K$ at line 12, and set $K' = \{u^- \neq v^- \mid u^- \neq v^- \in K\}$. By the construction of $K$ in lines $4 - 11$, all models of $Q'$ have to satisfy the formulae of $K - K'$. Further, using Theorem 5.9 and the construction of $K$, $H' \cup K'$ is also satisfiable, and thus $H' \cup K$ is satisfiable. Now, by the construction of $K$, line 13 only tests relations which have to hold in any model of $\langle V, E, H' \cup K \rangle$, and thus cannot reject, since then $Q'$ would not be satisfiable. Consequently, the algorithm accepts.

Suppose that $Q$ is not satisfiable, and that it accepts, meaning that neither of the tests at lines 2 or 13 succeed. At line 2, $Q'$ is not satisfiable, by Proposition 3.9. At line 4, $H'$ is satisfiable, and by the construction of $K$ in lines $4 - 11$, we have $def^+(Q', \langle V, E, H' \cup K \rangle)$ at line 12. By the same argument as above, $H' \cup K$ is satisfiable at line 12, using Theorem 5.9. If $\langle V, E, H' \cup K \rangle$ were locally satisfiable for starting points, then $\langle V, E, H' \cup K \rangle$ would be satisfiable, since $\mathcal{I}$ is a starting point algebra, and thus $Q'$ would be satisfiable, contrary to our assumption. Thus $\langle V, E, H' \cup K \rangle$ is not locally satisfiable.

Since the algorithm does not reject in line 14, the set

$$P = \{u^+ \epsilon prel^-(r)v^+ \mid \langle u, r, v \rangle \in E \wedge u^- = v^- \in H' \cup K\}$$

is satisfiable by some model $M$. We already know that there exists a model $N$ for $H' \cup K$. We shall construct an interpretation $M'$ from $M$ and $N$ which locally satisfies $\langle V, E, H' \cup K \rangle$, a contradiction.

Let $x$ be minimal such that for some $v \in V$, $x = M(v^+)$, i.e. the smallest ending point in $M$, and let $y$ be maximal such that for some $v \in V$, $y = N(v^-)$, i.e. the largest starting point in $N$. Now, for every $v \in V$, set $M'(v^-) = N(v^-)$ and $M'(v^+) = M(v^+) - x + y + 1$. We see that $M'$ still satisfies $P$, since the order on ending points is identical in $M$ and $M'$, and that it satisfies $H' \cup K$, since $M$ and $M'$ coincide on starting points. Furthermore, setting

$$E' = \{\langle u, r, v \rangle \in E \mid u^- = v^- \in H' \cup K\},$$

$\langle V, E', H' \cup K \rangle$ is satisfied in $M'$, since $H' \cup K$ and $P$ are, and since by construction $M(v^-) < M(v^+)$ is satisfied for every $v \in V$. But this is the same as $M'$ locally satisfying $\langle V, E, H' \cup K \rangle$, and this contradicts the fact that the algorithm does not reject. The result follows. $\square$

**Theorem 5.11** Algorithm 3.10 runs in polynomial time.

**Proof:** The transformation in line 1 is clearly polynomial and by Proposition 3.3, so is line 2. The loop between lines 5 to 11 is executed as many times as there are arcs in $E$, and lines 6 to 10 take only polynomial time, thus the loop takes polynomial time. The final test on line 12 is also polynomial-time (even linear), by Proposition 3.5, and the result follows. □

**Corollary 5.12** $M_s\text{-}ISAT(\mathcal{I})$ for starting point algebras, and $M_e\text{-}ISAT(\mathcal{I})$ for ending point algebras, is solvable in polynomial time.

**Proof:** From Theorem 5.10 and Theorem 5.11. □

For instances $Q = \langle V, E, H \rangle$ where $H$ only contains point algebra formulae, we can use PASAT instead of HORNDLRSAT in lines 2 and 6, since $expl^-(Q)$ adds only point algebra formulae to $H$ and $K$ in Algorithm 3.10 also contains only point algebra formulae. This allows for a much better worst-case complexity, as we shall see.

**Corollary 5.13** If Algorithm 3.10 uses PASAT instead of HORNDLRSAT in lines 2 and 6, its complexity becomes in $O(|E|^2)$.

**Proof:** The first transformation clearly takes $O(|E|)$ time. The initial and final tests take $O(|E|)$ time, by Proposition 3.5, and the loop is executed $|E|$ times, each taking $O(|E|)$ time. Thus, the resulting complexity is $O(|E|^2)$. □

Note that we do not have a proof that these algebras are the *only* algebras which are starting or ending point algebras in Allen's interval algebra. It seems likely that there should be more algebras with the same structure. One advantage of the results presented in this paper is that once a starting or ending point algebra is found (and proved to be one), both a polynomial-time algorithm for satisfiability and the extension to metric temporal information are obtained for free.

## 6. Maximality of Tractable Subclasses

Recently, the search for tractable subalgebras of Allen's interval algebra has become more systematic, by focusing on finding tractable algebras that are *maximal*, in the sense that no algebra strictly containing it is tractable. The pioneering work by Nebel and Bürckert (1995) was to find a maximal tractable subclass containing all the thirteen basic relations, the *ORD-Horn* algebra, which in addition is the *unique* maximal algebra containing all the basic relations. The ORD-Horn algebra contains 868 relations. More recently, Drakengren and Jonsson (1996) have identified nine more maximal tractable algebras, eight of which are of size 2178, and one of size 4097. However, the latter is found to be of no use, since any instance of satisfiability for that algebra is always satisfiable, unless it contains the relation $\perp$. The former algebras each contain three basic relations.

In Proposition 4.4 we saw that the twelve nonmaximal algebras of Drakengren and Jonsson (1996) are included in the algebras of the current paper. However, it remains to check whether they are maximal or not.

One of the main tools for analysing maximal tractability is a *closure operation* on subclasses of the algebra, which preserves tractability.

**Definition 6.1 (Closure)**

Let $S \subseteq \mathcal{A}$. Then we denote by $\overline{S}$ the *closure* of $S$ under converse, intersection and composition, i.e., the least subalgebra (which is uniquely determined) containing $S$, which is closed under the three operations. $\square$

The key result for extrapolating tractability results is the following.

**Proposition 6.2** Let $S \subseteq \mathcal{A}$. Then $ISAT(S)$ is polynomial-time iff $ISAT(\overline{S})$ is, and $ISAT(S)$ is NP-complete iff $ISAT(\overline{S})$ is.
**Proof:** See Nebel and Bürckert (1995). $\square$

Our main tools for proving intractability are the following NP-complete subalgebras of $\mathcal{A}$.

**Definition 6.3 (NP-complete algebras $\mathcal{N}_1$, $\mathcal{N}_2$ and $\Delta_0$)**

First let

$$A = \{(\prec \; \mathsf{d} \; \mathsf{o} \; \mathsf{m} \; \mathsf{f}^{\smile}), (\prec \; \mathsf{d} \; \mathsf{o} \; \mathsf{m} \; \mathsf{s})\},$$

define

$$\mathcal{N}_1 = A \cup \{(\mathsf{d} \; \mathsf{d}^{\smile} \; \mathsf{o}^{\smile} \; \mathsf{s}^{\smile} \; \mathsf{f})\}$$

and

$$\mathcal{N}_2 = A \cup \{(\mathsf{d}^{\smile} \; \mathsf{o} \; \mathsf{o}^{\smile} \; \mathsf{s}^{\smile} \; \mathsf{f}^{\smile})\}.$$

Also define

$$\Delta_0 = \{(\equiv \; \mathsf{d} \; \mathsf{d}^{\smile} \; \mathsf{o} \; \mathsf{o}^{\smile} \; \mathsf{m} \; \mathsf{m}^{\smile} \; \mathsf{s} \; \mathsf{s}^{\smile} \; \mathsf{f} \; \mathsf{f}^{\smile}), (\prec \; \succ)\}.$$

$\square$

**Proposition 6.4** $ISAT(\mathcal{N}_1)$, $ISAT(\mathcal{N}_2)$ and $ISAT(\Delta_0)$ are all NP-complete.
**Proof:** For $\mathcal{N}_1$ and $\mathcal{N}_2$, see Nebel and Bürckert (1995), and for $\Delta_0$, see Golumbic and Shamir (1993). $\square$

Next, we prove maximality of the current algebras.

**Proposition 6.5** The algebras $S(b)$, $E(b)$, $S^*$, $E^*$ are maximal tractable.
**Proof:** By running the utility $\mathtt{atry}$ (Nebel and Bürckert, 1993), which generates minimal extensions of subclasses by adding a relation and computing the closure of that class. All extensions of these algebras contain either of the NP-complete algebras $\mathcal{N}_1$, $\mathcal{N}_2$ or $\Delta_0$, so the result follows by Proposition 6.2. $\square$

Briefly, we show that the restriction that we cannot express starting and ending point information at the same time is essential for obtaining tractability, once we want to go outside the ORD-Horn algebra.

**Proposition 6.6** Let $S \subseteq \mathcal{A}$ such that $S$ is not a subset of the ORD-Horn algebra, and let $SE$ be the set of instances $Q = \langle V, E, H \rangle$ of $M$-$ISAT(S)$, where $H$ may contain only DLRs $u^+ = v^-$ for some $u, v \in V$. Then the satisfiability problem for $SE$ is NP-complete.
**Proof:** By definition, $S$ has to contain some relation outside the ORD-Horn algebra. Since the expression $u^+ = v^-$ is allowed in $H$, we can express the basic relation $\mathsf{m}$, and assume that it is included in $S$. It is easily verified that $\overline{\{\mathsf{m}\}}$ contains all the basic relations (using e.g. the utility `aclose` (Nebel and Bürckert, 1993)), and thus $\overline{\mathsf{s}}$ contains all basic relations. Since any subclass of Allen's algebra containing all the basic relations has to be contained in the ORD-Horn algebra in order to be tractable, and is NP-complete otherwise (Nebel and Bürckert, 1995), NP-completeness follows. $\square$

Despite this result, it seems possible to express constraints on *duration* of intervals, without restricting their absolute position in time, and still obtain a polynomial-time algorithm, but this is left for future work.

Now, one might question the relevance of this work on the grounds that this might be just eight out of thousands (or more) of tractable subalgebras. That is, we would like to have a result similar to the unique maximality result of the ORD-Horn class, showing that these algebras are the only algebras satisfying some specific criterion of relevance. In fact, recent results by the authors (Drakengren and Jonsson, 1997) state that any tractable subclass that is not yet known in the literature cannot contain more than three basic relation including converses (technically, it cannot contain basic relations other than ($\equiv$), ($b$) and ($b^\smile$) for $b \in \{\mathsf{d}, \mathsf{o}, \mathsf{s}, \mathsf{f}\}$). This means, for instance, that the two algebras $S(\succ)$ and $E(\prec)$ are the only maximal tractable algebras containing the relation $\prec$, and that any yet unpublished tractable subclass has to be less expressive in terms of the number of basic relations than the present algebras containing five basic relations.

## 7. Discussion

It seems appropriate to summarise the status of the search for maximal tractable subclasses of Allen's interval algebra: The eight new maximal tractable subalgebras presented in this paper increase the number of currently known maximal tractable subclasses to eighteen, including the ORD-Horn class (Nebel and Bürckert, 1995), and the nine algebras found by Drakengren and Jonsson (1996).

One may note that there is a considerable overlap between these, since the sizes of the algebras are 868 (one), 2178 (eight), 2312 (eight), and 4097 (one), the sum of the sizes being much more than 8192. For instance, we showed in Drakengren and Jonsson (1996) that the only relations of the ORD-Horn algebra that is not included in any of the algebras discussed in that paper are the relations ($\mathsf{m}$) and ($\mathsf{m}^\smile$).

Of course, the ultimate goal is to classify the *complete* set of maximal tractable subalgebras, but since there are $2^{8192}$ subclasses to investigate, this is clearly a nontrivial task. Recent results on the *RCC-5 algebra* for spatial reasoning (Jonsson and Drakengren, 1997) show that brute-force methods can have success in characterising the complete set of tractable subclasses. The present problem is harder with several orders of magnitude, however, since the RCC-5 algebra contains only $2^{32}$ relations. Nevertheless, it is encouraging to note that there are only four maximal tractable subclasses of the RCC-5 algebra, out of the approximately $4.3 \cdot 10^9$ subclasses. As mentioned above, recent results by the authors (Drakengren and Jonsson, 1997) also provide a partial classification of tractability in Allen's algebra, using similar methods.

Concerning metric time, it still remains to provide the nine maximal tractable algebras of Drakengren and Jonsson (1996) with some kind of metric temporal information. A simple examination shows that we cannot use the present technique, since this would make the resulting algebras NP-complete (just add the qualitative relations induced by relations on starting or ending points, run `aclose` (Nebel and Bürckert, 1993), and verify that at least one of the NP-complete algebras of Proposition 6.4 is included), so for this, some other kind of expressivity is needed. Searching for other starting or ending point algebras could also be fruitful. Further, it seems possible that the techniques presented here can also be used for extending the point-interval algebra (Vilain, 1982) with metric time.

## 8. Conclusions

We have found eight new maximal tractable subclasses of Allen's interval algebra, and provided them with metric temporal information on starting or ending points of intervals, using the formalism of Horn DLRs (Jonsson and Bäckström, 1996). Apart from representing progress in the research aiming at a complete characterisation of the tractable subclasses of Allen's interval algebra, this opens for a combination between the expressivity of the ORD-Horn algebra and algebras which can express *sequentiality* between intervals, provided that only starting or ending points of intervals are related with ORD-Horn relations.

## Acknowledgements

Thanks to Christer Bäckström, Henry Kautz, and the two anonymous reviewers for helpful comments.